\documentclass[runningheads]{llncs}
\usepackage{amsfonts}
\usepackage[T1]{fontenc}
\usepackage{graphicx}
\usepackage{amsmath}
\usepackage{multirow}

\begin{document}
\title{CIBR: Cross-modal Information Bottleneck Regularization for Robust CLIP Generalization}
\author{
\begin{tabular}{c}
\textbf{Yingrui Ji}$^{1,7}$ \quad
\textbf{Xi Xiao}$^{2}$ \quad
\textbf{Gaofei Chen}$^{2}$ \quad
\textbf{Hao Xu}$^{3}$ \\
\textbf{Chenrui Ma}$^{6}$ \quad
\textbf{Lijing Zhu}$^{5}$ \quad
\textbf{Aokun Liang}$^{4}$ \quad
\textbf{Jiansheng Chen}$^{1\dagger}$
\end{tabular}
}

\institute{
$^1$ Aerospace Information Research Institute, Chinese Academy of Sciences \\
$^2$ University of Alabama at Birmingham \\
$^3$ Harvard Medical School \\
$^4$ Wuhan University \\
$^5$ Bowling Green State University \\
$^6$ University of California, Irvine \\
$^7$ School of Electronic, Electrical and Communication Engineering, University of Chinese Academy of Sciences\\
$^\dagger$ Corresponding author}

\maketitle              % typeset the header of the contribution
\pagestyle{plain} 
\vspace{-2em}
\begin{abstract}

Contrastive Language-Image Pretraining (CLIP) has achieved remarkable success in cross-modal tasks such as zero-shot image classification and text–image retrieval by effectively aligning visual and textual representations. However, the theoretical foundations underlying CLIP’s strong generalization remain unclear. In this work, we address this gap by proposing the Cross-modal Information Bottleneck (CIB) framework. CIB offers a principled interpretation of CLIP’s contrastive learning objective as an implicit Information Bottleneck optimization. Under this view, the model maximizes shared cross-modal information while discarding modality-specific redundancies, thereby preserving essential semantic alignment across modalities. Building on this insight, we introduce a Cross-modal Information Bottleneck Regularization (CIBR) method that explicitly enforces these IB principles during training. CIBR introduces a penalty term to discourage modality-specific redundancy, thereby enhancing semantic alignment between image and text features. We validate CIBR on extensive vision–language benchmarks, including zero-shot classification across seven diverse image datasets and text–image retrieval on MSCOCO and Flickr30K. The results show consistent performance gains over standard CLIP. These findings provide the first theoretical understanding of CLIP’s generalization through the IB lens. They also demonstrate practical improvements, offering guidance for future cross-modal representation learning.

\end{abstract}
\section{Introduction}
\vspace{-0.8em}
Contrastive Language-Image Pretraining (CLIP)~\cite{radford2021learning} has demonstrated significant progress in multimodal representation learning, enabling effective generalization across diverse vision-language tasks. Unlike traditional supervised visual models, CLIP employs large-scale, noisy image-text datasets and leverages contrastive learning to align visual and textual embeddings. This training paradigm results in remarkable zero-shot capabilities, facilitating transfer to a wide range of downstream tasks without task-specific fine-tuning~\cite{radford2021learning,zhou2022learning}. Despite these empirical successes, the underlying theoretical reasons for CLIP’s strong cross-modal generalization remain insufficiently understood. Current research primarily aims at enhancing CLIP through empical methods, including sophisticated prompt engineering  and advanced fine-tuning strategies~\cite{zhou2022learning,khattak2023maple,bahng2022visual}, yet these approaches provide limited theoretical insights. The critical question remains unresolved:\textit{ what theoretical mechanism enables CLIP’s strong generalization capabilities across modalities?}

To address this gap, we revisit CLIP through the lens of Information Bottleneck (IB) theory~\cite{tishby2000information,tishby2015deep}, aiming to fundamentally explain CLIP's extraordinary generalization capability. IB theory offers a rigorous framework for understanding generalization by positing that effective representations compress input data while retaining task-relevant information~\cite{shwartz2017opening,saxe2019information,ying2025Enhancing}. We hypothesize that CLIP implicitly achieves an IB-like optimization, simultaneously compressing irrelevant, modality-specific redundancies and preserving essential cross-modal semantic correlations. In this paper, we formalize this intuition by introducing a novel theoretical framework, named Cross-modal Information Bottleneck (CIB), explicitly linking CLIP's contrastive learning objectives to IB principles.
\begin{figure}[htbp]
\vspace{-1em}
    \centering
    \includegraphics[width=0.6\linewidth]{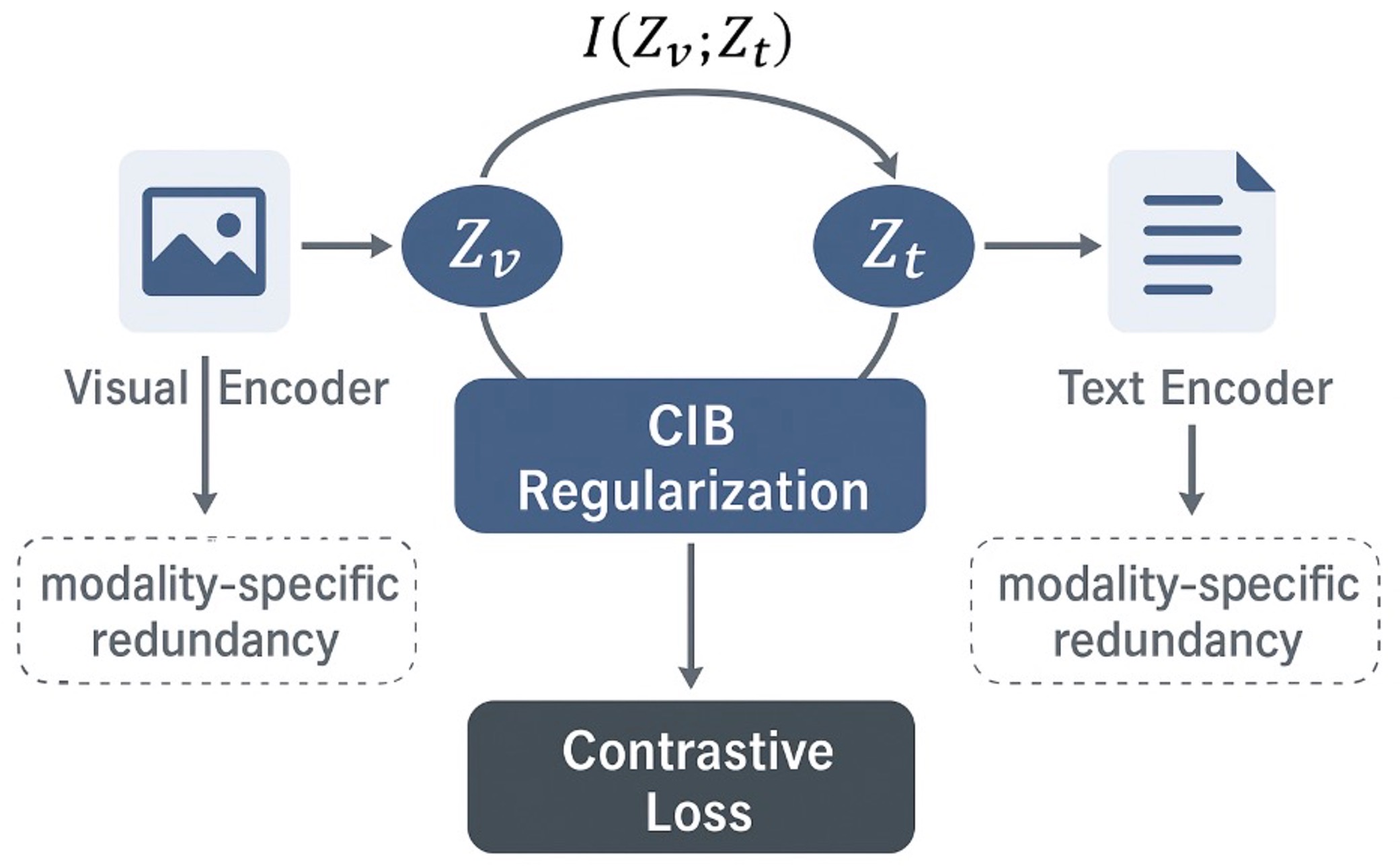}
    \caption{Illustration of our Cross-modal Information Bottleneck Regularization (CIBR) strategy. Given image-text pairs, the visual and textual encoders produce embeddings $Z_v$ and $Z_t$, which are aligned through contrastive learning. }
    \label{fig:intro_cibr}
    \vspace{-1em}
\end{figure}
\begin{itemize}
    \item We provide the first comprehensive theoretical interpretation of CLIP from the Information Bottleneck perspective, elucidating the intrinsic reasons behind its strong cross-modal generalization.
    \item Based on IB theory, we propose a novel mutual-information regularization strategy, explicitly guiding the CLIP model towards optimal cross-modal representation learning.
    \item Our theoretical insights are validated through preliminary experiments on representative multimodal datasets, highlighting the effectiveness and practical value of theory-informed approaches in contrastive multimodal learning.
\end{itemize}

\section{Related Work}

\subsection{Contrastive Language-Image Pretraining (CLIP)}
Contrastive Language-Image Pretraining (CLIP)~\cite{radford2021learning,zhou2022learning,fang2023eva,yao2023filip} has emerged as a significant breakthrough in multimodal representation learning by effectively bridging vision and language modalities. Trained on millions of noisy image-text pairs, CLIP optimizes a contrastive objective that aligns representations from the two modalities into a unified embedding space. This enables zero-shot transfer learning across numerous downstream tasks, significantly outperforming traditional supervised methods on various vision benchmarks~\cite{radford2021learning,zhou2022learning}. Despite its empirical success, current studies mainly concentrate on further empirical improvements, such as CoOp~\cite{zhou2022conditional} and MaPLe~\cite{khattak2023maple}, and ALIGN~\cite{jia2021scaling}, which explore effective prompting mechanisms and large-scale multimodal alignment. Moreover, LiT~\cite{zhai2022lit} further enhances the transferability by leveraging pre-trained vision and text models. Despite these empirical successes, theoretical understanding of CLIP’s generalization capabilities across modalities remains under-explored. While some recent theoretical works examine the representational geometry~\cite{wang2022chaos} or learning dynamics~\cite{saunshi2022understanding} in unimodal contrastive settings, rigorous analyses specifically addressing CLIP's cross-modal generalization remain scarce. Understanding the underlying theoretical mechanisms driving CLIP’s remarkable zero-shot capabilities thus represents an important yet open research direction.
\cite{radford2021learning,zhou2022learning}. Despite its empirical success, current studies mainly concentrate on further empirical improvements. Moreover, LiT~\cite{zhai2022lit} enhances transferability by leveraging pre-trained vision and text models. Nonetheless, theoretical understanding of CLIP’s cross-modal generalization remains under-explored. While recent theoretical works examine representational geometry~\cite{wang2022chaos} or learning dynamics~\cite{saunshi2022understanding} in unimodal contrastive settings, rigorous analyses explicitly addressing CLIP's cross-modal generalization remain scarce. Therefore, uncovering the theoretical foundations behind CLIP’s remarkable zero-shot capabilities is an important yet open research direction.

\subsection{Prompt Tuning in Vision Models}
Prompt tuning originated in the Natural Language Processing (NLP) domain as a parameter-efficient fine-tuning strategy, allowing large pretrained models to adapt quickly to downstream tasks by learning additional soft prompts~\cite{li2021prefix}. Prompt tuning was soon adapted to vision foundation models, notably Vision Transformers (ViT)~\cite{dosovitskiy2021an,jia2022visual,xiao2025tdrdtopdownbenchmarkrealtime}. For instance, Visual Prompt Tuning (VPT)~\cite{jia2022visual} introduced learnable prompts to the input and intermediate feature spaces of ViT, achieving promising performance with minimal parameter updates. Recent advances further refine prompt tuning by incorporating semantic priors, such as textures, shapes, and color histograms, significantly enhancing generalization and interpretability~\cite{bahng2022visual,han2023e2vpt,xiao2025visualvariationalautoencoderprompt}. More recent works, such as Multi-modal Prompt Learning~\cite{khattak2023maple} and Cross-modal Prompt Tuning~\cite{wang2023multimodal}, have demonstrated the effectiveness of prompt-based methods in cross-modal learning scenarios. However, these methods predominantly remain empirically driven, with limited theoretical insights guiding the design and selection of prompts in the visual domain.

\subsection{Information Bottleneck in Deep Learning}
The IB theory, first introduced by Tishby~\cite{tishby2000information} and ~\cite{shwartz2017opening,tishby2015deep}, provides a principled framework for understanding representation learning in neural networks. The IB principle aims to find the optimal representation by compressing the input data, preserving only information relevant to target labels.~\cite{shwartz2017opening,tishby2015deep}. Specifically, IB theory explains the generalization capabilities of neural networks through the lens of mutual information between input, representations, and task labels~\cite{belghazi2018mutual,saxe2019information}. Goldfeld~\cite{goldfeld2020information} and Alemi~\cite{alemi2016deep} further extend these theoretical analyses by providing formal connections between neural network optimization and IB principles. Nonetheless, existing IB-driven analyses have mainly focused on single-modality data, lacking rigorous extension to cross-modal scenarios, especially for contrastive learning frameworks such as CLIP.

\subsection{Theoretical Analysis of Contrastive Learning}
Contrastive learning has attracted substantial theoretical interest due to its impressive empirical performance in unsupervised and self-supervised learning scenarios~\cite{chen2020simple,he2020momentum}. Recent theoretical works primarily focus on deriving generalization bounds using the Rademacher complexity and PAC-Bayes frameworks, highlighting how contrastive objectives implicitly optimize mutual information between different augmentations of the same instance~\cite{wang2021understanding,haochen2021provable,ji2024deep}. Notably, Saunshi et al.~\cite{saunshi2022understanding} provided insightful theoretical analysis demonstrating when and why contrastive objectives yield strong representations. Nevertheless, these theoretical analyses primarily emphasize unimodal settings (such as vision or NLP alone), and theoretical understanding of cross-modal contrastive learning—especially in models like CLIP—remains inadequate.

\subsection{Cross-modal Representation Learning}
Cross-modal representation learning has rapidly advanced by learning unified embeddings across modalities~\cite{alayrac2020self,qi2020imagebert}. Several empirical methods have been proposed, including cross-modal alignment~\cite{jia2021scaling}, cross-modal retrieval~\cite{wang2021continual}, and modality fusion~\cite{di2021align}. Despite the remarkable empirical progress, a comprehensive theoretical understanding of cross-modal alignment mechanisms and their impacts on model generalization remains scarce~\cite{tsai2022conditional,wang2023connecting}.

\section{Cross-modal Information Bottleneck Analysis}

\begin{figure}[htbp]
\vspace{-1.5em}
    \centering
    \includegraphics[width=\linewidth]{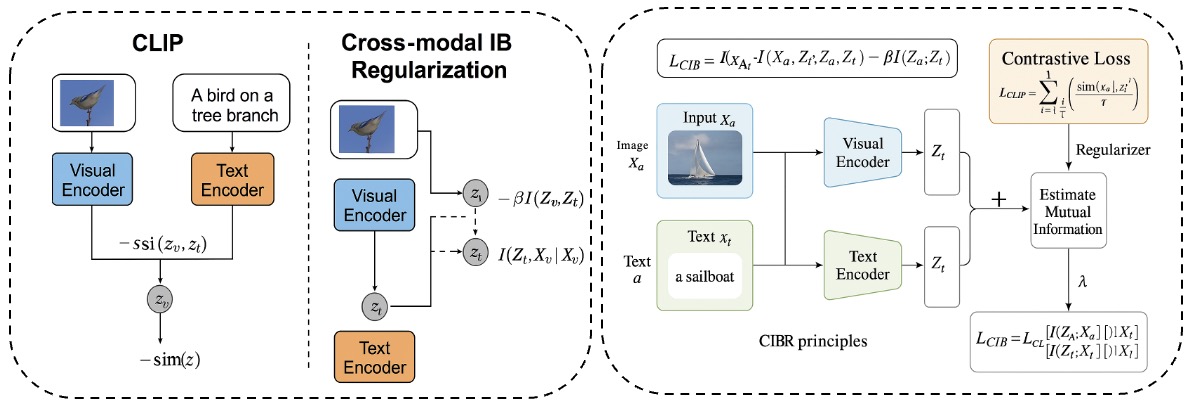}
    \caption{Comparison between the original CLIP architecture and our proposed CIBR framework. The left diagram shows the standard CLIP structure that aligns image-text pairs via contrastive loss. The right diagram illustrates our enhanced pipeline, where an information bottleneck-inspired regularization module explicitly suppresses modality-specific redundancy and improves cross-modal semantic alignment by maximizing the mutual information $I(Z_v; Z_t)$ while minimizing conditional redundancy.}
    \label{fig:framework}
\end{figure}

In this section, we first revisit the IB principle~\cite{tishby2000information,tishby2015deep} and then propose our novel theoretical framework, termed CIB, to rigorously analyze how CLIP implicitly performs a cross-modal information optimization.  To clearly illustrate our proposed learning pipeline, we present the overall architecture of the CIBR model in Figure~\ref{fig:framework}.

\vspace{-1em}
\subsection{Information Bottleneck Principle}
The IB principle, initially proposed by Tishby ~\cite{tishby2000information}, provides a rigorous theoretical framework for analyzing deep neural network training from an information-theoretic perspective. Formally, given an input random variable $X$ and a target variable $Y$, the IB principle seeks a compact representation $Z$ that balances the trade-off between compressing input redundancy and preserving relevant information for the target task. This is achieved by minimizing the following IB objective:
\begin{equation}
    \mathcal{L}_{IB} = I(X;Z) - \beta I(Z;Y),
\end{equation}
where $X$, $Z$, and $Y$ denote the input, representation, and task label, respectively, and $\beta$ controls the trade-off between compression and relevance~\cite{tishby2015deep,shwartz2017opening}.

\subsection{Cross-modal Information Bottleneck Framework}
Inspired by IB theory, we extend it to the multimodal context and define the CIB framework. Specifically, consider the two modalities involved in CLIP training: visual modality $X_v$ and textual modality $X_t$. The representation learning objective in cross-modal tasks seeks to learn joint representations $Z_v$ and $Z_t$ that preserve semantic-relevant information while discarding modality-specific redundancies. Formally, we define the cross-modal IB objective as:
\begin{equation}
    \mathcal{L}_{CIB} = I(X_v,X_t;Z_v,Z_t) - \beta I(Z_v;Z_t),
\end{equation}
where $I(\cdot;\cdot)$ represents mutual information between random variables. Minimizing $\mathcal{L}_{IB}$ implicitly enforces the model to encode cross-modal correlations while compressing irrelevant modality-specific information.

\subsection{Connecting CLIP to Cross-modal IB}
We now theoretically analyze how the contrastive loss function used by CLIP implicitly aligns with the proposed cross-modal IB framework. The CLIP contrastive loss~\cite{radford2021learning} can be formulated as:
\begin{equation}
    \mathcal{L}_{CLIP} = -\frac{1}{N}\sum_{i=1}^{N}\log\frac{\exp(\text{sim}(z_v^i,z_t^i)/\tau)}{\sum_{j=1}^{N}\exp(\text{sim}(z_v^i,z_t^j)/\tau)},
\end{equation}
where $z_v^i$ and $z_t^i$ are visual and textual embeddings, respectively, and $\tau$ is a temperature parameter. This contrastive loss encourages high similarity between matched modality pairs and low similarity otherwise, effectively maximizing cross-modal mutual information $I(Z_v; Z_t)$. Simultaneously, due to the competitive nature of contrastive learning, irrelevant modality-specific features that do not contribute directly to cross-modal alignment are implicitly suppressed, thereby achieving the IB principle's compression objective. To formalize this intuition, we provide the following theoretical interpretation:

\textbf{Proposition 1:} \textit{Minimizing the CLIP contrastive loss implicitly approximates the optimization of the cross-modal information bottleneck objective, where the representation learned by CLIP attempts to maximize $I(Z_v;Z_t)$ and minimize modality-specific redundancies, i.e., compressing $I(X_v; Z_v|X_t)$ and $I(X_t;Z_t|X_v)$.} \textit{(Proof sketch)}: The contrastive loss can be viewed as maximizing cross-modal mutual information $I(Z_v;Z_t)$~\cite{oord2018representation,tschannen2019mutual}. Due to finite representational capacity and regularization inherent in optimization (such as batch size constraints, limited embedding dimensions, and noise from stochastic gradient descent), irrelevant modality-specific redundancy (noise) is implicitly reduced. Thus, optimizing the contrastive loss inherently aligns with the proposed CIB objective.

\subsection{Cross-modal IB-inspired Regularization}
Based on the above theoretical insights, we propose an explicit regularization scheme named \textbf{CIBR} to strengthen CLIP’s adherence to IB constraints. Specifically, we introduce an explicit regularizer based on mutual information estimation techniques~\cite{belghazi2018mutual} to encourage further compression of modality-specific redundancies and better semantic alignment:
\begin{equation}
    \mathcal{L}_{CIBR} = \mathcal{L}_{CLIP} + \lambda [I(Z_v;X_v|X_t) + I(Z_t;X_t|X_v)],
\end{equation}
where $\lambda$ balances the strength of regularization. Intuitively, this additional regularizer explicitly penalizes modality-specific redundancies and thus theoretically enhances generalization.

% \subsection{Theoretical Implications and Discussion}
% Our theoretical analysis offers critical insights into CLIP's strong generalization properties: CLIP implicitly performs cross-modal IB optimization by simultaneously aligning semantic correlations across modalities and discarding irrelevant modality-specific information. Explicitly enforcing IB constraints (via CIBR) can theoretically lead to better generalization by minimizing redundancy and emphasizing semantic alignment. 

\section{Cross-modal Information Bottleneck Regularization}
Building on the theoretical insights above, we propose a regularization strategy—\textbf{CIBR}—to explicitly improve the cross-modal generalization of CLIP-based models. While CLIP implicitly optimizes a cross-modal IB objective during contrastive training, we argue that explicitly enforcing IB constraints can further enhance semantic alignment and robustness. Motivated by this, we introduce an IB-inspired regularizer based on mutual information estimation~\cite{belghazi2018mutual}, which encourages the model to compress modality-specific redundancies while maximizing the semantic correlation between visual and textual representations. In the following, we detail its formulation, theoretical basis, and practical integration.

\subsection{CIB Regularization Formulation}

As discussed in Section 3.2, the optimal cross-modal representation learning can be formally expressed by minimizing the Cross-modal Information Bottleneck objective in Eq.(2). Direct computation of mutual information terms is challenging in practice, especially for multimodal high-dimensional spaces. Thus, we derive a practical approximation to this objective based on mutual information neural estimation techniques.
% \begin{equation}
%     \mathcal{L}_{CIB} = I(X_v, X_t; Z_v, Z_t) - \beta I(Z_v, Z_t; Y),
% \end{equation}
% where $X_v$, $X_t$ represent visual and textual inputs, respectively; $Z_v$, $Z_t$ denote the corresponding learned representations, and $Y$ represents semantic labels or cross-modal alignment labels. Direct computation of mutual information terms is challenging in practice, especially for multimodal high-dimensional spaces. Thus, we derive a practical approximation to this objective based on mutual information neural estimation techniques.

\subsection{Mutual Information Estimation in Cross-modal Embedding}
To effectively implement our IB-inspired regularization, accurate estimation of mutual information is crucial. Following~\cite{belghazi2018mutual}, we adopt the Mutual Information Neural Estimator (MINE) to estimate mutual information between modality-specific and cross-modal embeddings. MINE estimates mutual information through neural networks, defined as follows:
\begin{equation}
    I_\theta(X; Z) = \sup_{\theta\in\Theta}\mathbb{E}_{p(x,z)}[T_\theta(x,z)] - \log(\mathbb{E}_{p(x)p(z)}[e^{T(x,z)}]),
\end{equation}
where $T_\theta$ is a neural network parameterized by $\theta$. Through gradient-based optimization, we iteratively update $T_\theta$ to obtain precise mutual information estimates.

\subsection{CIBR Optimization Objective}

With MINE-based mutual information estimation, we formally define our proposed Cross-modal Information Bottleneck Regularization as an additional regularization term appended to the original CLIP contrastive loss. The complete optimization objective becomes:

\begin{equation}
\mathcal{L}_{CIBR} = \mathcal{L}_{CLIP} + \lambda \left[I(Z_v;X_v|X_t) + I(Z_t;X_t|X_v)\right],
\end{equation}
where $I(Z_v;X_v|X_t)$ and $I(Z_t;X_t|X_v)$ represent conditional mutual information terms that quantify modality-specific redundant information. Intuitively, minimizing these terms explicitly penalizes the retention of redundant modality-specific information not contributing directly to cross-modal semantic alignment, thus theoretically improving the generalization ability of learned embeddings.

In practice, we approximate these conditional mutual information terms via MINE, optimizing the regularization term simultaneously alongside CLIP’s contrastive objective during training. The combined loss function can be expressed as:
\begin{equation}
    \mathcal{L}_{total} = \mathcal{L}_{CLIP} + \lambda\left(I(Z_v;X_v|X_t) + I(Z_t;X_t|X_v)\right),
\end{equation}
where $\lambda$ is a hyperparameter controlling the strength of the regularization. Empirically, tuning $\lambda$ allows us to balance cross-modal alignment accuracy and redundancy reduction, effectively optimizing the model’s IB trade-off.

% \subsection{Implementation Details}
% We implement the proposed regularization strategy upon the standard CLIP architecture~\cite{radford2021learning}, keeping its visual and textual encoders unchanged. Specifically, we introduce two lightweight neural networks $T_v$ and $T_t$ to approximate mutual information terms through MINE. These networks take modality-specific embeddings and cross-modal representations as inputs and produce scalar outputs estimating corresponding mutual information terms. During training, parameters of $T_v$ and $T_t$ are optimized simultaneously with the standard CLIP model parameters through an iterative mini-batch gradient descent approach.

% \subsection{Theoretical Justifications}
% Our regularization explicitly guides the learning process toward representations satisfying the IB constraints, thus theoretically ensuring improved generalization. To demonstrate the effectiveness of CIBR from a theoretical standpoint, we argue that explicitly penalizing modality-specific redundancy inherently leads to better-conditioned optimization landscapes and tighter generalization bounds, following insights from recent studies~\cite{tishby2015deep,saxe2019information}. More formally, enforcing IB constraints directly restricts the representation complexity, thus reducing the effective hypothesis space complexity, aligning with known PAC-Bayes generalization results~\cite{saxe2019information}. Hence, our method is theoretically expected to exhibit superior generalization across unseen modalities compared to standard CLIP training.

\section{Experiment}

We evaluate the proposed CIBR on a broad set of vision-language benchmarks, including zero-shot image classification and text-image retrieval, following the CLIP evaluation protocol~\cite{radford2021learning}. No task-specific fine-tuning is used; class labels are embedded using handcrafted prompts as in the original setup. All models use ViT-B/32 as the visual backbone. We compare CIBR with CLIP, CoOp~\cite{zhou2022conditional}, and MaPLe~\cite{khattak2023maple}. For classification, we test on seven datasets: ImageNet, CIFAR-100, OxfordPets, Caltech101, DTD, SUN397, and EuroSAT—covering general object recognition, low-resolution images, fine-grained categories, texture, scenes, and remote sensing. As shown in Table~\ref{tab:zeroshot-full}, CIBR consistently improves accuracy across all benchmarks, with notable gains on fine-grained and texture-sensitive datasets. For example, it outperforms MaPLe by 1.5\% on OxfordPets and 1.3\% on DTD, confirming that enforcing IB constraints helps focus on semantically relevant features.
\begin{figure}[htbp]
\vspace{-1.5em}
    \centering
    \includegraphics[width=0.6\linewidth]{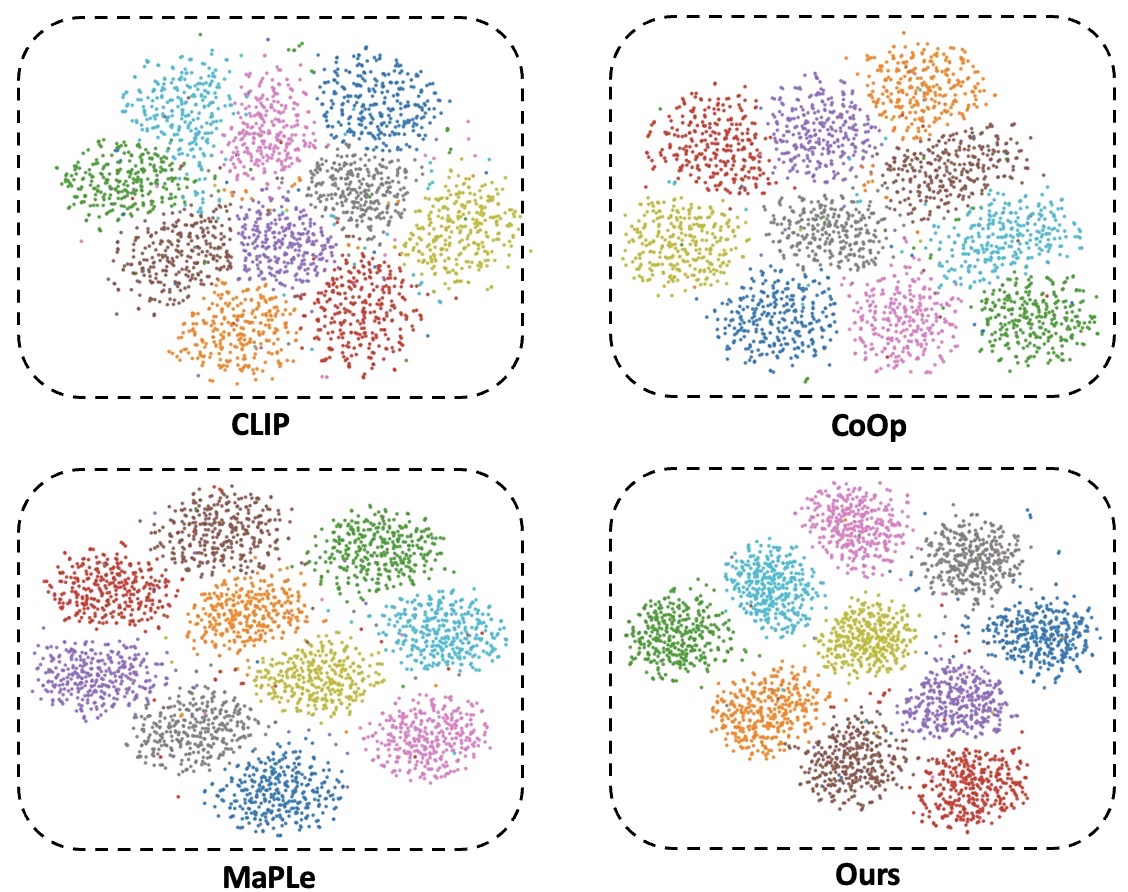}
    \caption{t-SNE visualization of learned feature embeddings on the CIFAR-100 dataset. }
    \label{fig:tsne_cifar100}
\end{figure}
\begin{table}[htbp]
\vspace{-2em}
\centering
\caption{Zero-shot classification accuracy (\%) across 7 benchmark datasets.}
\label{tab:zeroshot-full}
\begin{tabular}{l|ccccccc}
\hline
\textbf{Method} & ImageNet & CIFAR-100 & OxfordPets & Caltech101 & DTD & SUN397 & EuroSAT \\
\hline
CLIP & 63.3 & 71.2 & 88.3 & 92.5 & 72.3 & 60.5 & 63.1 \\
CoOp & 65.6 & 74.5 & 89.2 & 93.6 & 73.5 & 62.7 & 66.8 \\
MaPLe & 65.7 & 76.1 & 89.5 & 94.0 & 74.2 & 63.3 & 67.1 \\
CIBR (Ours) & \textbf{66.2} & \textbf{76.8} & \textbf{91.0} & \textbf{94.4} & \textbf{75.5} & \textbf{64.1} & \textbf{68.0} \\
\hline
\end{tabular}
\vspace{-2em}
\end{table}

To assess the model’s capability in aligning textual and visual modalities, we perform cross-modal retrieval on the MSCOCO and Flickr30K datasets, which are widely adopted in prior work. We report Recall@1, Recall@5, and Recall@10 for text-to-image retrieval. As shown in Table~\ref{tab:retrieval-full}, our model significantly outperforms the baselines across all retrieval metrics. On MSCOCO, CIBR achieves 63.4\% Recall@1, improving over CLIP by 4.8 points. On Flickr30K, which is more fine-grained and linguistically challenging, the advantage is even more evident, highlighting the benefit of removing modality-specific noise while enhancing semantic alignment.

\begin{table}[htbp]
\vspace{-1.5em}
\centering
\caption{Text-to-image retrieval performance (Recall@\{1,5,10\}) on MSCOCO and Flickr30K.}
\label{tab:retrieval-full}
\begin{tabular}{l|ccc|ccc}
\hline
\multirow{2}{*}{\textbf{Method}} & \multicolumn{3}{c|}{MSCOCO} & \multicolumn{3}{c}{Flickr30K} \\
 & R@1 & R@5 & R@10 & R@1 & R@5 & R@10 \\
\hline
CLIP & 58.6 & 83.0 & 89.5 & 63.2 & 86.4 & 91.8 \\
CoOp & 60.5 & 84.3 & 90.2 & 64.0 & 87.2 & 92.5 \\
MaPLe & 61.7 & 85.1 & 91.1 & 65.1 & 88.0 & 93.0 \\
CIBR (Ours) & \textbf{63.4} & \textbf{86.8} & \textbf{92.3} & \textbf{66.7} & \textbf{89.1} & \textbf{94.1} \\
\hline
\end{tabular}
\vspace{-1.5em}
\end{table}

\begin{figure}[htbp]
    \centering
    \begin{minipage}[t]{0.49\linewidth}
        \centering
        \includegraphics[width=\linewidth]{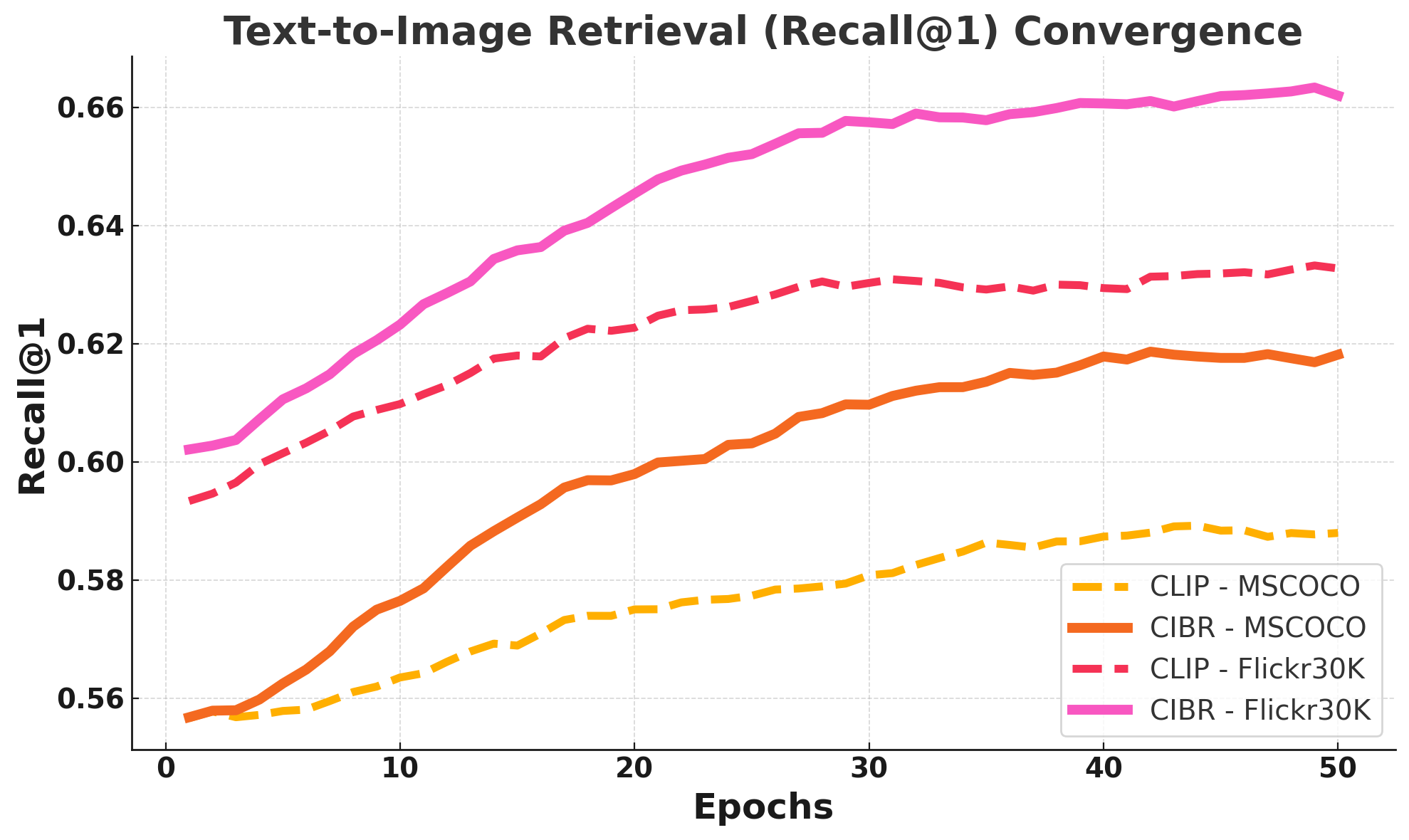}
        \caption{Text-to-image retrieval performance (Recall@1) during training on MSCOCO and Flickr30K. }
        \label{fig:recall_convergence}
    \end{minipage}
    \hfill
    \begin{minipage}[t]{0.49\linewidth}
        \centering
        \includegraphics[width=\linewidth]{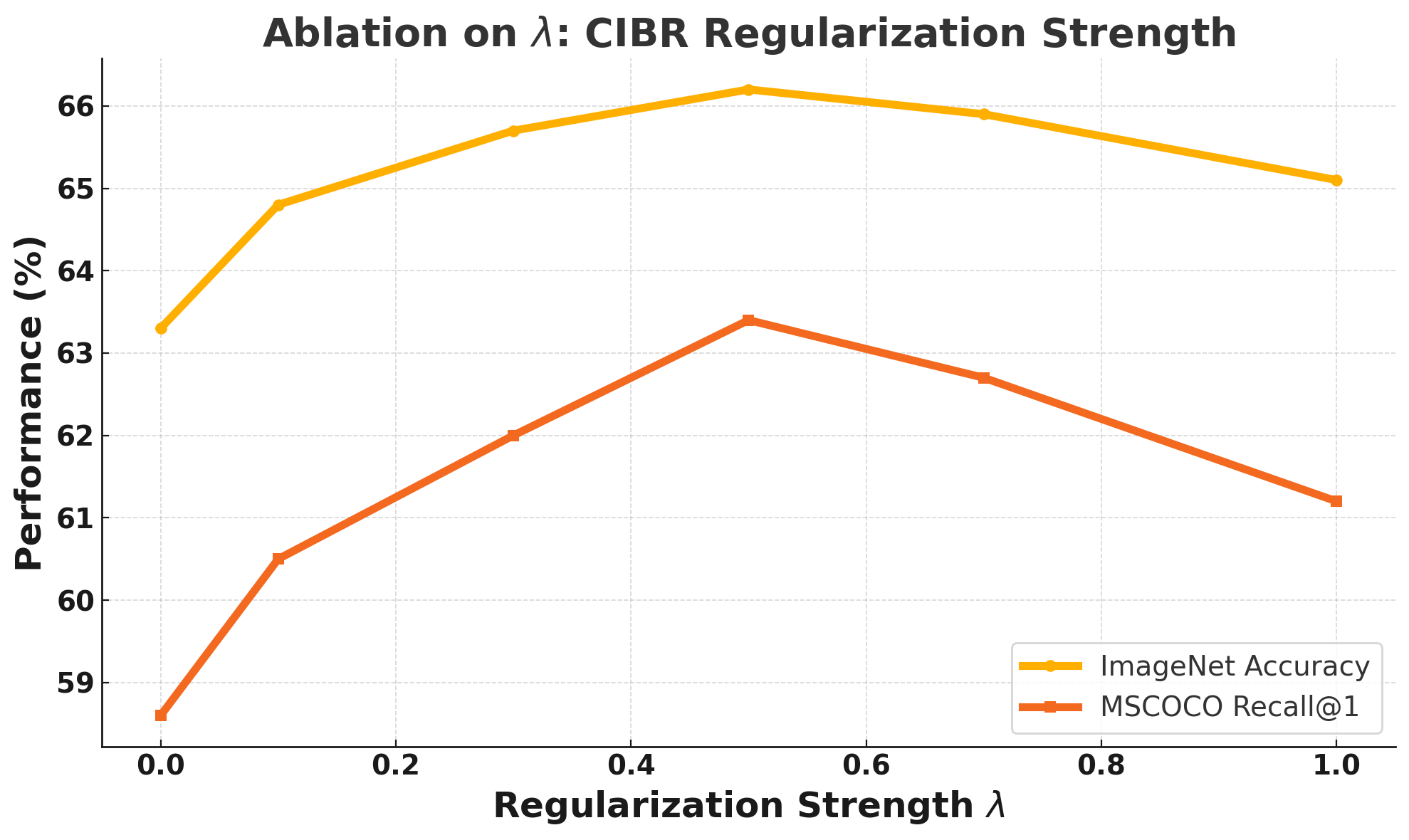}
        \caption{Ablation study on the regularization coefficient $\lambda$ in the CIBR loss. }
        \label{fig:lambda_ablation}
    \end{minipage}
    \vspace{-1.5em}
\end{figure}

We also investigate the influence of the regularization strength $\lambda$ on the model’s performance. An ablation on ImageNet reveals that increasing $\lambda$ improves accuracy up to a point, beyond which excessive regularization slightly degrades performance. This observation aligns with the theoretical trade-off in information bottleneck theory: too little regularization fails to compress irrelevant features, while too much suppresses informative content. Figure~\ref{fig:lambda_ablation} illustrates this effect, showing a peak around $\lambda=0.5$.

\subsection{Visualization and Analysis of Training Dynamics}

\begin{figure}[htbp]

    \centering
    \begin{minipage}[t]{0.49\linewidth}
        \centering
        \includegraphics[width=\linewidth]{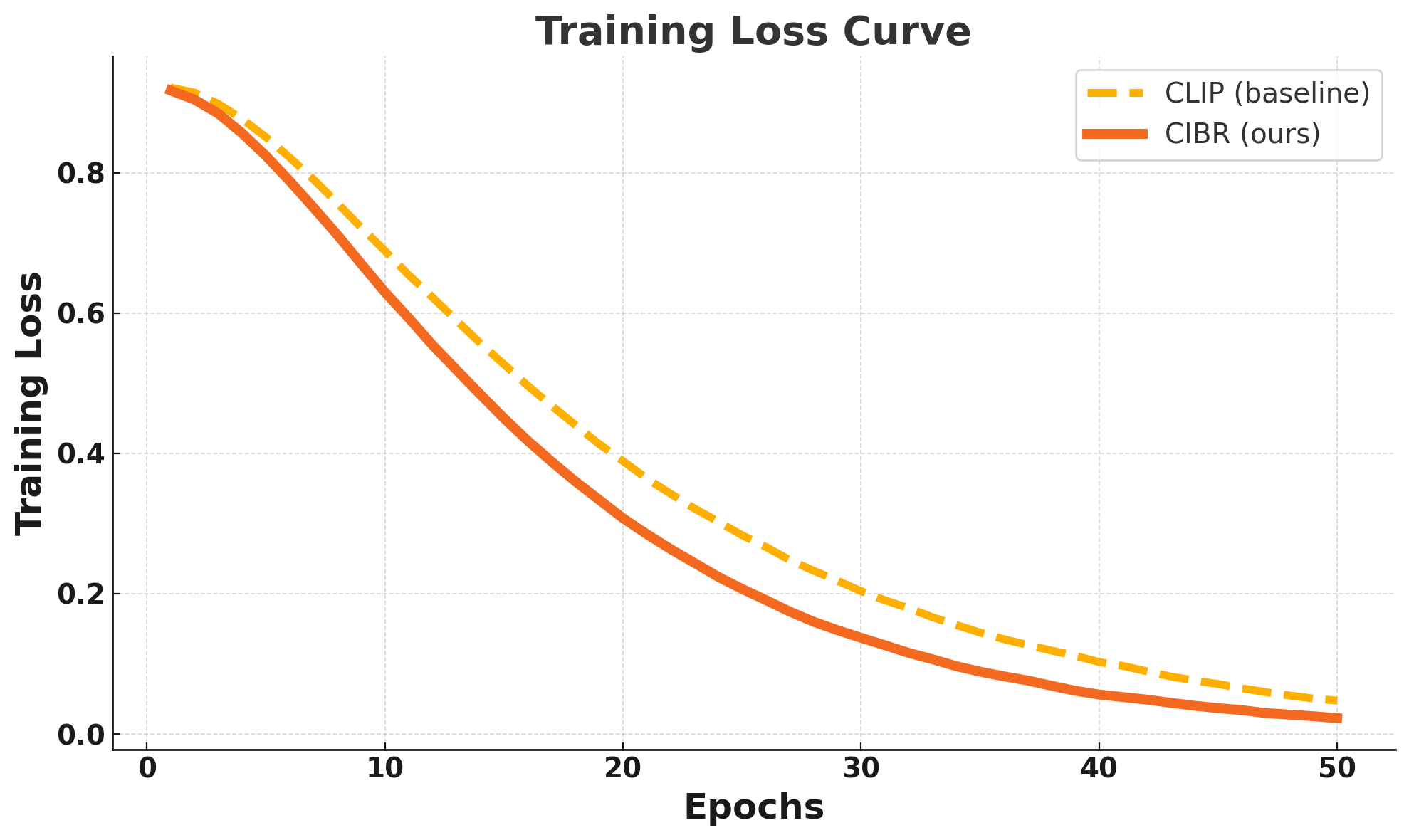}
        \caption{Training loss curves of CLIP and our proposed CIBR method on the training set.}
        \label{fig:training_loss}
    \end{minipage}
    \hfill
    \begin{minipage}[t]{0.49\linewidth}
        \centering
        \includegraphics[width=\linewidth]{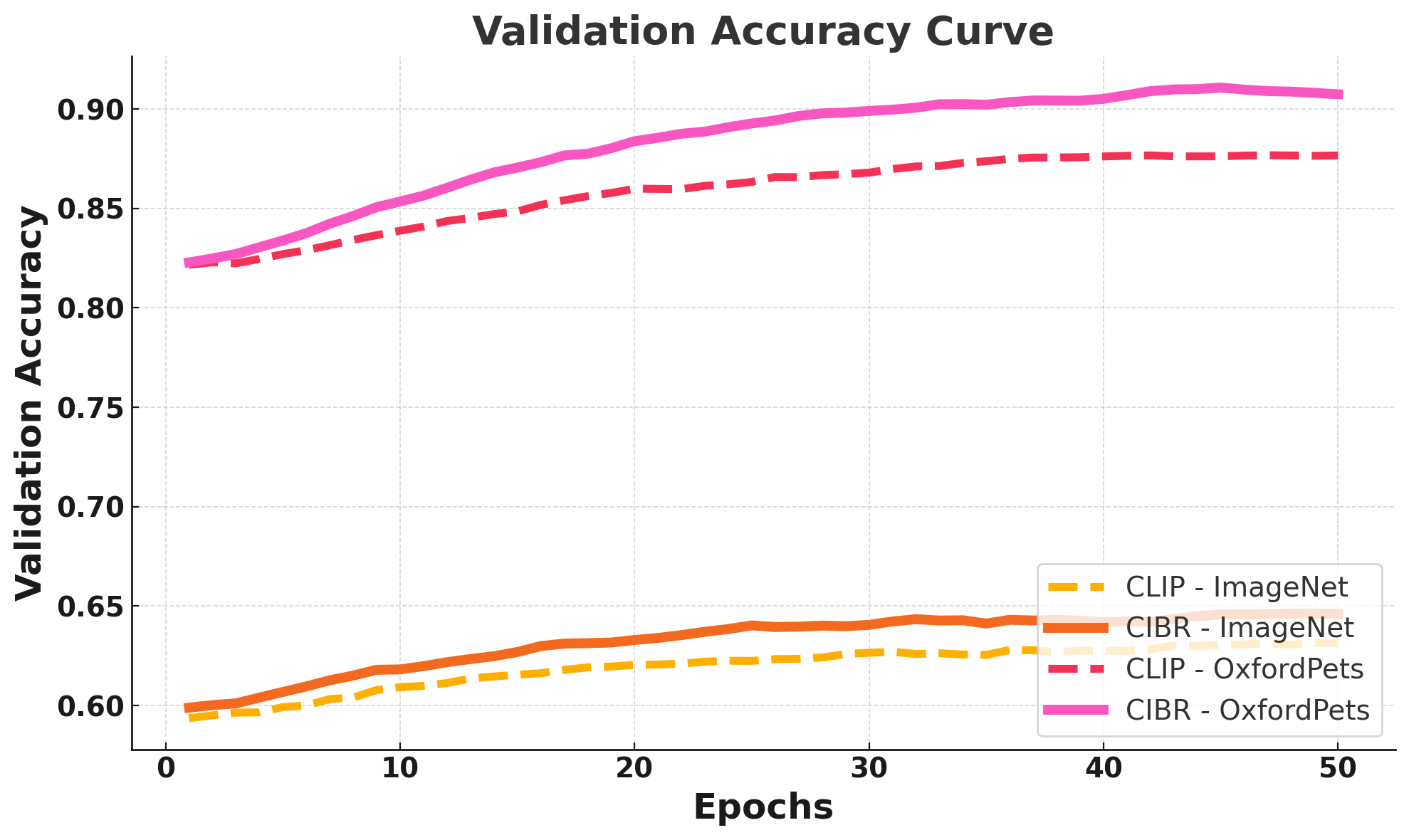}
        \caption{Validation accuracy curves of CLIP and CIBR on ImageNet and OxfordPets. }
        \label{fig:val_accuracy}
    \end{minipage}
    \vspace{-1.2em}
\end{figure}

\begin{figure}[htbp]
    \centering
    \begin{minipage}[t]{0.49\linewidth}
    \includegraphics[width=\linewidth]{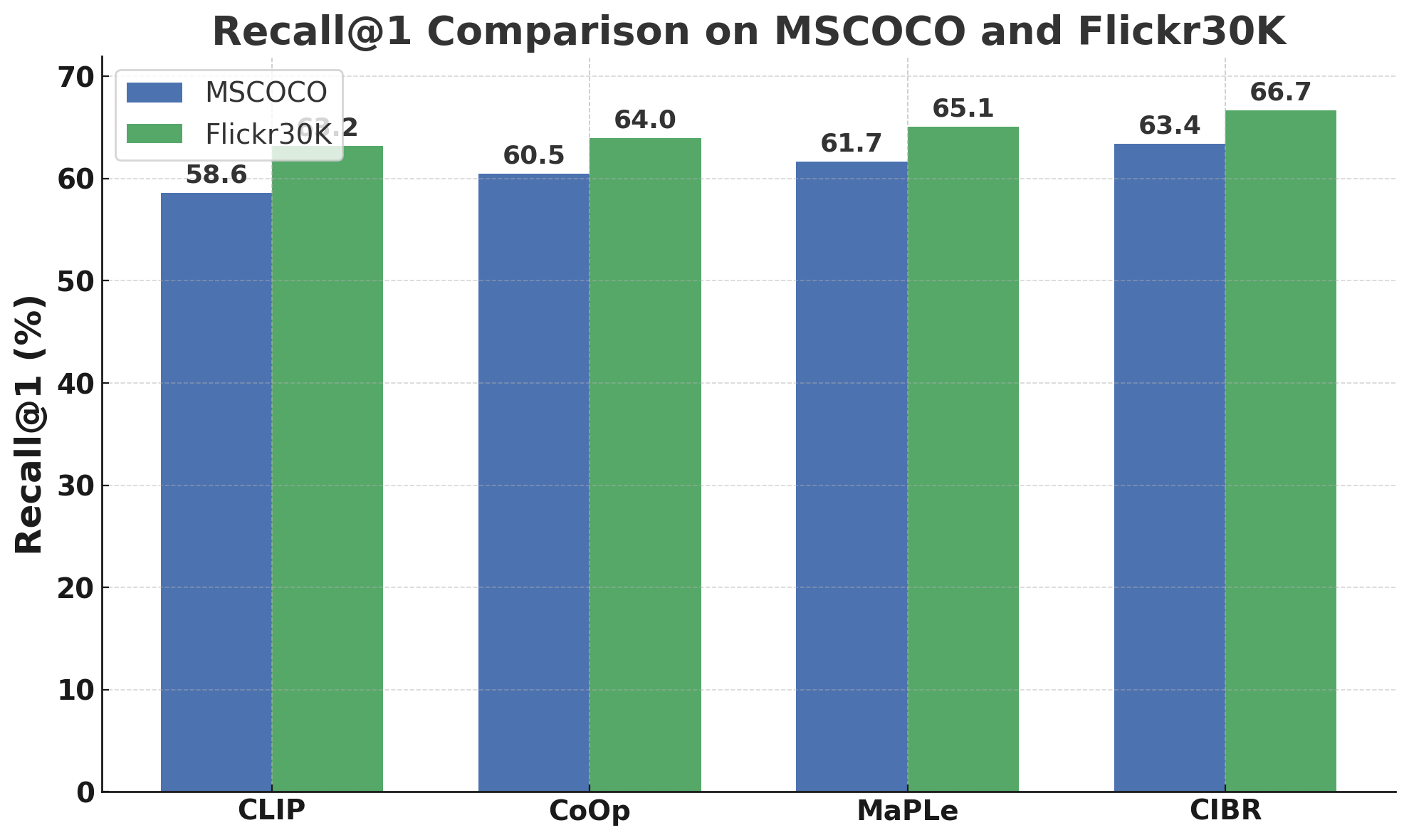}
    \caption{Final Recall@1 performance comparison across four methods (CLIP, CoOp, MaPLe, CIBR) on MSCOCO and Flickr30K. }
    \label{fig:recall_bar_chart}
    \end{minipage}
    \hfill
    \begin{minipage}[t]{0.49\linewidth}
    \includegraphics[width=\linewidth]{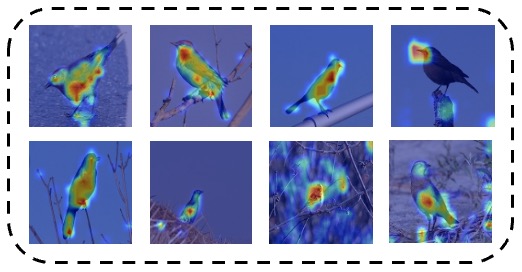}
    \caption{Grad-CAM visualization results from our proposed model on bird classification samples. }
    \label{fig:gradcam_samples}
    \end{minipage}
    \vspace{-2em}
\end{figure}

We present visualizations to analyze how CIBR improves optimization, representation quality, and interpretability. \textbf{Training and Retrieval Dynamics.} As shown in Figures~\ref{fig:training_loss} and~\ref{fig:val_accuracy}, CIBR converges faster and achieves lower training loss and higher validation accuracy than CLIP, indicating more stable optimization and better generalization. Retrieval results in Figure~\ref{fig:recall_convergence} show improved Recall@1 and convergence speed on MSCOCO and Flickr30K. Figure~\ref{fig:recall_bar_chart} further confirms CIBR’s superiority over CLIP, CoOp, and MaPLe in final retrieval performance. \textbf{Regularization Strength.} Figure~\ref{fig:lambda_ablation} shows performance varies with the regularization weight $\lambda$. A moderate value ($\lambda \approx 0.5$) yields the best results, consistent with the theoretical trade-off in IB. \textbf{Interpretability and Feature Structure.} Grad-CAM visualizations (Figure~\ref{fig:gradcam_samples}) show CIBR attends more accurately to semantically meaningful regions, improving interpretability. t-SNE plots (Figure~\ref{fig:tsne_cifar100}) illustrate tighter intra-class clusters and better inter-class separation, indicating enhanced feature discrimination and semantic alignment. These results validate that CIBR improves not only task performance but also the structure and interpretability of cross-modal representations.

% In summary, our experiments demonstrate that the proposed CIBR framework consistently enhances the generalization capability of CLIP across a wide range of datasets and tasks. These improvements are especially notable in challenging fine-grained or domain-shift scenarios, validating the theoretical intuition that cross-modal generalization benefits from discarding redundant modality-specific information. Moreover, CIBR introduces minimal architectural overhead and is compatible with existing contrastive training pipelines, making it both effective and practical for real-world deployment.

\section{Discussion and Conclusion}

This paper presents a theoretical reinterpretation of CLIP from the perspective of IB theory. We introduce the CIB framework, which reveals that CLIP implicitly maximizes cross-modal mutual information while compressing modality-specific redundancies—offering a principled explanation for its strong generalization. Building on this insight, we propose CIBR, which explicitly enforces IB constraints via mutual information estimation. Extensive experiments on zero-shot classification and cross-modal retrieval tasks confirm that CIBR improves performance while enhancing representation quality. Despite promising results, future work will focus on more scalable mutual information estimation and broader empirical validation across diverse multimodal benchmarks. Overall, our work bridges theoretical analysis and practical improvement, providing a solid foundation for principled cross-modal learning.


\bibliographystyle{splncs04}
\bibliography{sample-base}
%





% LNCS Homepage, \url{http://www.springer.com/lncs}, last accessed 2023/10/25
% \end{thebibliography}
\end{document}